\documentclass[10pt,leqno]{article}
\usepackage[utf8]{inputenc}
\usepackage[T1]{fontenc}
\usepackage{authblk}            
\usepackage{graphicx,amssymb,amsmath,amsthm,csquotes}
\usepackage{xcolor,paralist,hyperref,titlesec,fancyhdr,etoolbox}
\usepackage[margin=1in]{geometry}
\usepackage{lipsum}

\hypersetup{colorlinks=true,linkcolor=black,filecolor=black,urlcolor=black}

\titleformat{\section}[display]{\normalfont\huge\bfseries\centering}{\thesection}{10pt}{\Large}
\titlespacing*{\section}{0pt}{0ex}{0ex}

\usepackage{array} 
\newcolumntype{C}[1]{>{\centering\arraybackslash}m{#1}}

\title{ORCHID: Orchestrated Retrieval-Augmented Classification with Human-in-the-Loop Intelligent Decision-Making for High-Risk Property}

\newcommand{\equalcontrib}{\textsuperscript{*}}

\author[1]{Maria Mahbub\equalcontrib}
\author[1]{Vanessa Lama\equalcontrib}
\author[1]{Sanjay Das}
\author[2]{Brian Starks}
\author[1]{Christopher Polchek}
\author[1]{Saffell Silvers}
\author[1]{Lauren Deck}
\author[1]{Prasanna Balaprakash}
\author[1]{Tirthankar Ghosal}

\affil[1]{Oak Ridge National Laboratory, Oak Ridge, TN, USA}
\affil[2]{Pacific Northwest National Laboratory, Richland, WA, USA}

\date{} 

\begin{document}
\maketitle

\begingroup
  \renewcommand\thefootnote{\textsuperscript{*}}
  \footnotetext{Both authors contributed equally to this research.}
\endgroup

\begin{abstract}
High-Risk Property (HRP) classification is critical at U.S. Department of Energy (DOE) sites, where inventories include sensitive and often dual-use equipment. Compliance must track evolving rules designated by various export control policies to make transparent and auditable decisions. Traditional expert-only workflows are time-consuming, backlog-prone, and struggle to keep pace with shifting regulatory boundaries.
We demo ORCHID, a modular agentic system for HRP classification that pairs retrieval-augmented generation (RAG) with human oversight to produce policy based outputs that can be audited. Small cooperating agents—retrieval, description refiner, classifier, validator, and feedback logger—coordinate via agent-to-agent messaging and invoke tools through the Model Context Protocol (MCP) for model-agnostic on-premise operation. The interface follows an Item to Evidence to Decision loop with step-by-step reasoning, on-policy citations, and append-only audit bundles (run-cards, prompts, evidence). In preliminary tests on real HRP cases, ORCHID improves accuracy and traceability over a non-agentic baseline while deferring uncertain items to Subject Matter Experts (SMEs). The demonstration shows single item submission, grounded citations, SME feedback capture, and exportable audit artifacts—illustrating a practical path to trustworthy LLM assistance in sensitive DOE compliance workflows.
\end{abstract}


\flushbottom
\maketitle

\thispagestyle{empty}
\section*{Introduction}
Early efforts to automate export‐control and security classification relied on rules and ontologies curated by subject-matter experts (sme), typically wrapping the eCFR United States Munitions List (USML) \footnote{https://www.ecfr.gov/current/title-22/chapter-I/subchapter-M/part-121}, eCFR Nuclear Regulatory Commission (NRC) \footnote{https://www.ecfr.gov/current/title-10/chapter-I}, and the eCFR Commerce Control List (CCL) \footnote{https://www.ecfr.gov/current/title-15/subtitle-B/chapter-VII/subchapter-C/part-774} into machine-readable taxonomies\cite{li2020ontology, clark2008automated}. These systems improved consistency but struggled with ambiguous cross-category items and frequent rule changes. Recent research has moved toward knowledge-centered and ontology-driven modeling of security/export-control concepts, enabling richer reasoning over product descriptions and technical attributes \cite{rao2009meta}. For example, ontology-based security/export-control classification approaches demonstrate that standardized concept graphs can reduce ambiguity and support explainable labeling \cite{rafaki25}, though coverage gaps remain for rapidly evolving technologies (e.g., advanced semiconductors, dual-use AI) \cite{dualAI}.

In parallel, legal-domain NLP benchmarks and pipelines have matured, offering reusable evaluation settings for statutory retrieval, classification, and entailment. LEXGLUE aggregates legal tasks (e.g., multi-label classification, case entailment) and established baselines, while LegalBench focuses on statute understanding and legal reasoning with LLMs \cite{chalkidis2021lexglue}.

Retrieval-augmented generation (RAG) has become a dominant strategy for keeping models aligned to authoritative texts and reducing hallucinations in law and governance applications \cite{reuter2025reliableretrievalragsystems, Huang_2025}. Industry and academic reports alike emphasize dynamic retrieval from up-to-date regulatory repositories and explicit citation in output. Legal-tech guidance and empirical frameworks such as “dynamic legal RAG,”\cite{ajay2025optimizing} Gov-RAG\cite{yu2025gov}, and SemRAG \cite{zhong2025semrag} all report improvements in factuality and traceability when generation is grounded in statutes, regulatory notices, and agency FAQs. These capabilities are essential for export-control determinations.

Recent research in AI has explored various methods for combining human expertise with machine learning models. Active learning and human-in-the-loop systems have been widely used in tasks where expert feedback can help refine models, especially in high-stakes domains like medical diagnostics and security \cite{razkaz25, WANG2024103201}. Additionally, retrieval-augmented models like RAG have shown promise in tasks that require both context-specific information retrieval and generation, such as legal document review and scientific research assistance.
However, few systems focus on integrating SME feedback in real-time to refine predictions. Furthermore, the domain of classifying high-risk properties for national labs requires both high accuracy and clear explanations of predictions, which presents unique challenges for AI models. Our work aims to address this gap by using a human-in-the-loop approach combined with RAG to improve the classification process.
\section*{Design   for ORCHID}
\label{sec:orchid-design}
ORCHID supports real-word decision context backed by a verifiable system guaranty. The design focuses on providing a traceable decision pattern in the world of generative AI supported by on-policy citations, step-by-step reasoning for High-risk Property (HRP) determination, and incorporated SME feedback; all of which is written in an immutable audit log. 

\subsection*{Task Setting \& Stakeholders} 
ORCHID allows personnel to analyze the HRP status of items during procurement. Items, based on their HRP status, are flagged by the tool and forwarded to SMEs for review if they do not pass a certain threshold of confidence score set by the stakeholders. SMEs then provide a thorough review of the item(s) and provide their feedback that is cached into a database and tagged to the item for future instances of using the tool to classify a "similar" item. 

During procurement, items are required to be categorized as high-risk or not. Items controlled under the instructions of the USML and NRC regulations are considered high-risk level 1; level 1 being the "top-priority" items to be regulated as per regulatory requirements. Those that are controlled under the CCL are considered high-risk level 2. CCL items are referred to as ``dual-usage'' items due to their commercial and military applications. All items that aren't in the high-risk categories are deemed to be low-risk or labelled as EAR99.

Thus, ORCHID design prioritizes stakeholder's desire for a solution that tags each item's HPR status along with an HRP list categorization (USML, NRC, CCL), i.e. if the item(s) are determined to be high-risk. Another equal priority is that the solution provides a traceable and a natural language representation of the decision process for the classification. As input, the manufacturer, the equipment or service, and the model number are available along with an optional user description of the item. 
\subsection*{Principles} 
ORCHID's design is driven by the need for a traceability-first, policy-aware approach to solve the classification problem with the integration of human-in-the-loop. We tackle the problem in two-folds, retrieval, and classification.

\textit{Evidence-first:} Every classification must be tagged with at least one citation to USML, NRC, or CCL text. Any conflicting snippets are also visible.

\textit{Human-in-the-loop:} First model outputs are considered proposals. Thus, if the proposals have low model confidence, a reviewer must Accept/Override the item with certain feedback.

\textit{Grounded retrieval only:} Search runs over a whitelisted and versioned policy corpus using hybrid retrieval (BM25 + embeddings + RRF).
    
\textit{Reproducible by default:} Each session writes an audit bundle of inputs, outputs, and every decision pattern that can be logged within the tool.

\textit{Secure on-prem:} No external data sharing in in-prem system. Logs are only for governance and not for training/fine-tuning.

\textit{Configurable and fail-safe:} Thresholds can be enabled to exercise a cut-off for "send to SME" decision. When inputs are ambiguous, ORCHID defers to human.

These rules shape how information moves through ORCHID. As we explore the architecture of the system, we'll also learn how different portions of the tool interact. 

\subsection*{Out-of-Scope and Safety Guardrails} 
The demo artifact keeps scope tight and privacy guarantees explicit. It does not auto-approve any decision: every user-visible outcome is gated by the validator and, when flagged, requires a human click with a short rationale. It does not hide conflicting evidence; contradictory snippets remain visible so reviewers can understand edge cases. Retrieval does not mix in out-of-scope sources—only the versioned USML/NRC/CCL/EAR99 corpus is queried, \textbf{with web search disabled}. In on-prem mode there is no external egress: hosted model routes are off and outbound calls are blocked; the interface makes this state visible. Collected artifacts are not reused for training or fine-tuning and audit bundles exist only for governance. Finally, the demo avoids irreversible actions without confirmation and logging, and focuses on a single-item flow (bulk queues are possible by running scripts but does not currently have a user interface).

\section*{The ORCHID System} 
\label{sec:orchid-system}
\subsection*{Agentic Workflow} 
\begin{figure}[!t]
    \centering
    \includegraphics[width=\linewidth]{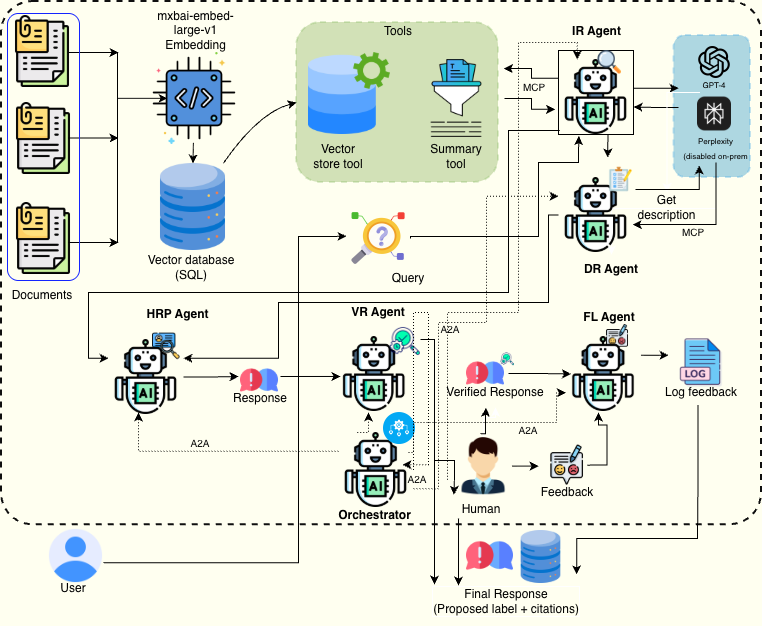}
    \caption{ORCHID agentic workflow.
The Orchestrator coordinates agents for Retrieval (IR), Description Refinement (DR), HRP classification, Validation (VR), and Feedback Logging (FL) via agent-to-agent (A2A) messages. IR/DR/HRP access local tools through Model Context Protocol (MCP) adapters (Vector Store, Summary) over a versioned policy corpus. VR either issues a verified decision or routes the case to a human reviewer, whose feedback is recorded in an append-only audit log.}
    \label{fig:agentic-arch}
\end{figure}

ORCHID runs a small set of cooperating agents behind a thin controller (“Orchestrator”). The controller publishes/consumes typed events on a local message bus and enforces contracts so each step is reproducible: item IDs, config hash, model versions, and timestamps travel with every message. In our implementation, agentic batch run script creates the session, sends submit to the bus, and the Orchestrator schedules agents in order, handling retries and timeouts. The response aggregator normalizes partial output, making each stage safe to run and tolerant of absent keys. The following further description into each of the agents as portrayed in Fig.~\ref{fig:agentic-arch}.
    
\textit{IR (Information Retrieval)} turns the operator’s fields (Manufacturer, Equipment/Service, Model No., optional description) into hybrid queries and calls the Vector Store tool to fetch policy-scoped snippets.

\textit{DR (Description Refiner)} is a helper that asks the operator for a clearer description or rewrites the provided text; it never calls cloud services in on-prem mode.

\textit{HRP (Classifier)} assembles the grounded prompt (fields + citations) and proposes a provisional decision: label + confidence + cited snippets.

\textit{VR (Validator)} checks coverage (enough on-policy support) and conflict (contradictory snippets), emitting a lightweight verdict between "AGREE", "REVIEW ", or "CONFLICT". VR is the gate; it either emits the Final Response or routes to a human.

\textit{FL (Feedback Logger)} records reviewer decisions and a one-line rationale into the append-only audit store.

\textit{Orchestrator} schedules the above, passes context between them, and annotates every event with the run-card (retrieval/model versions and parameters). It never generates model content.

Agent-to-Agent (A2A) is our internal message schema: type, item\_id, state, context, citations, decision, validator, and provenance. Agents read/write only these fields and anything missing has default set by the controller so a dictionary of list is never an assumption..

Model Context Protocol (MCP) is the way agents call local tools in a uniform way. We expose the Vector Store tool (hybrid search, re-rank, fuse) and a Summary tool (snippet condenser) as MCP adapters. On-prem, these are local processes with JSON I/O and data never leaves the system. IR/DR/HRP call them through the same contract, so swapping implementations is a configuration change, not a code change. Because every step depends on trustworthy evidence, we describe how retrieval and grounding work next.

\subsection*{Retrieval \& Grounding} 

ORCHID restricts search to a versioned policy corpus (eCFR USML, NRC, CCL, plus EAR99 guidance). Policy text is chunked with stable section IDs and embedded with \textit{mxbai-embed-large-v1}\cite{emb2024mxbai}\cite{li2023angle}. We maintain a hybrid index: BM25 over normalized text for lexical matches and a vector index for semantic matches and results are combined with reciprocal rank fusion. The Vector Store tool (MCP) encapsulates this. Agents pass a query object (with/without description, top-k, weights), and receive a ranked list of snippets with section IDs, confidence, and offsets. A small citation packer filters to minimally sufficient spans that the model must cite verbatim.

Grounding is enforced in two ways. First, prompts are constructed from item fields and packed snippets. The classifier cannot see free web text or out-of-scope sources. Second, the validator measures coverage (are there enough on-policy snippets supporting the label?) and conflict (do contradictory snippets appear?), producing the AGREE/REVIEW/CONFLICT signal that gates the final response. Operators can toggle whether the optional description is used. Retrieval falls back to fields-only when the description is noisy. All artifacts—queries, snippet IDs, prompts, outputs, verdicts—are stored with a run-card so the same decision can be replayed.
\newline
With evidence under control, we outline the system planes and API surface that drive single-item and batch runs.

\subsection*{User Interface}
\begin{figure*}[!htbp]
    \centering
    \includegraphics[width=\linewidth]{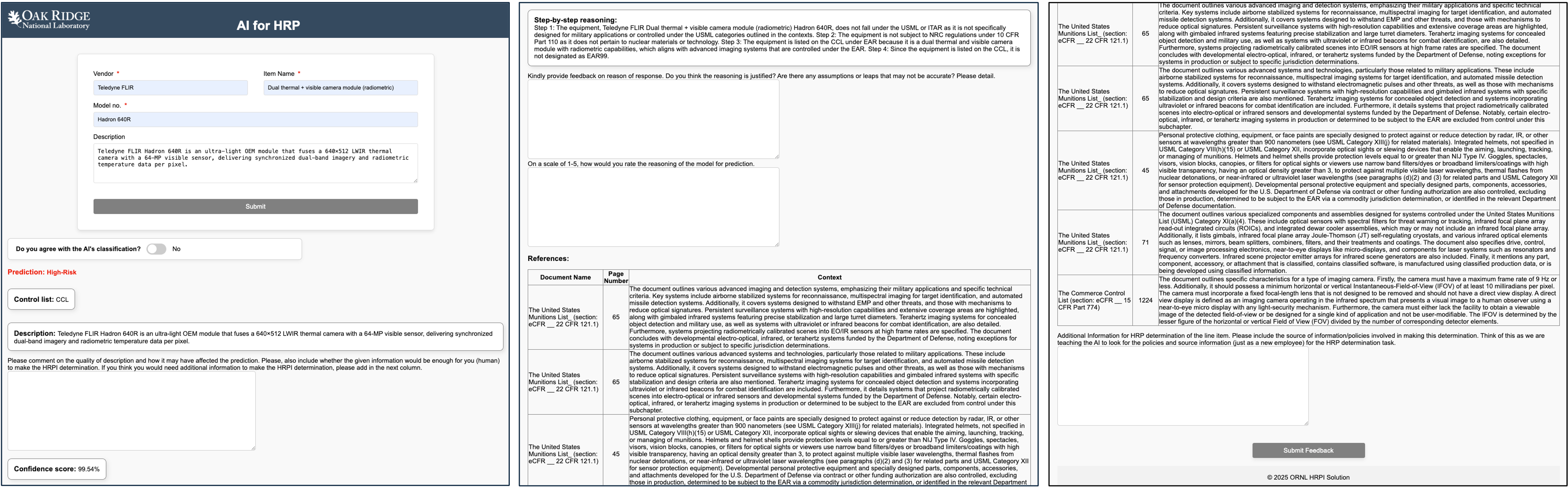}
    \caption{ORCHID UI overview.
Submit (vendor, item, model, optional description), inspect policy evidence with citations, review the proposed label and confidence, then record SME feedback}
    \label{fig:ui}
\end{figure*}

ORCHID's user interface is designed to streamline human–AI interaction in all stages of classification, focusing on transparency, traceability, and minimal analyst effort.
After setup, users are directed to the main interface where submissions, results, and feedback are managed through a unified workflow, as visible in Fig.~\ref{fig:ui}.

\textit{Submission Workflow:} Users initiate the classification workflow by entering a vendor name, item name, and model number, with an optional description, and then trigger processing with a one-click action via the `Submit' button. This workflow supports both individual entries and batch uploads for multiple items.

\textit{Outputs:} Upon submission, the interface displays the system’s prediction (HRP or Not HRP), the predicted control category, and a single confidence score summarizing model certainty.

\textit{Reasoning and Evidence:} Each result includes a concise, step-by-step reasoning trace supported by clickable citations. An evidence table presents the underlying documents, sections or pages, and extracted text, with actions for quick copy or open, as well as trace IDs for provenance tracking.

\textit{Feedback and Review:} Users can provide structured feedback -- agreement status, notes, rating, or policy reference -- through a simple form submitted via the `Submit Feedback' button. This input is stored for audit and model refinement.

\textit{Batch Mode and Export:} For large-scale reviews, the same interface supports batch processing with per-item status indicators and downloadable results. Completed analyses can be exported in JSON, CSV, or PDF formats, each embedding a version strip that records the model identifier, index snapshot, and timestamp to ensure auditability.

The application’s frontend provides a cohesive, context-aware experience, guiding users seamlessly from submission to reasoning review, feedback, and export, ensuring both operational efficiency and traceable decision support.

\section*{Demo Scenarios \& Walkthrough} 
\label{sec:orchid-demo}

The demo video shows ORCHID running on-prem on low-sensitivity, synthetic data. To build the example, we asked a GPT-5 chatbot to generate a random procurement item plus a claimed ground truth; for the video, the ground truth is CCL. The goal is to illustrate how the agentic workflow turns item information into a verified decision with citations while also incorporating a human reviewer. No external data sharing occurs in this demo: (Demo video will be available upon publication).
\newline

\section*{Preliminary Results}\label{sec:orchid-results}

\begin{figure}[!htbp]
  \centering
  \includegraphics[width=0.8\columnwidth]{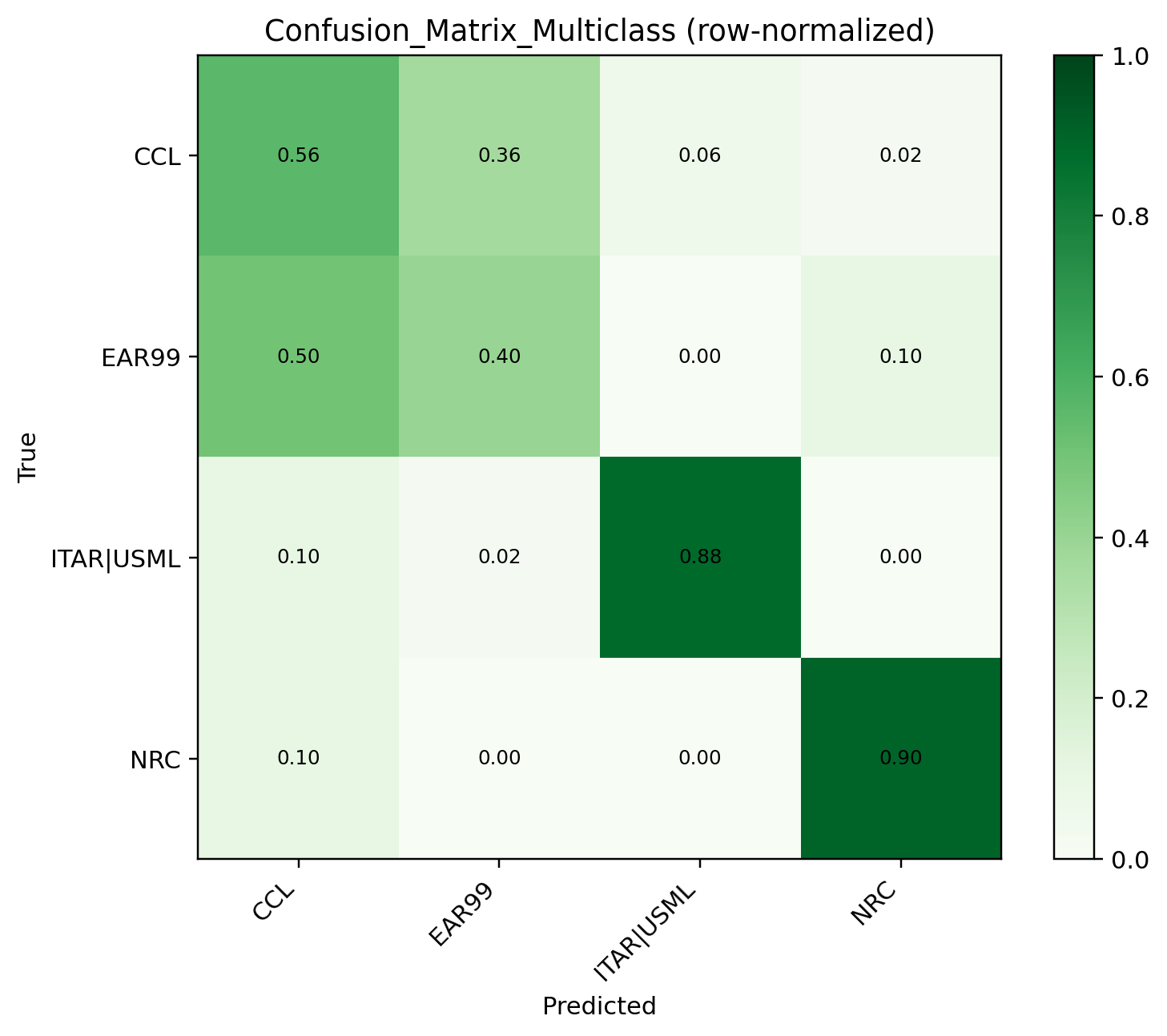}
  \caption{Heatmap with true classes on the y-axis and predicted classes on the x-axis (USML, NRC, CCL, EAR99); values are row-normalized.}
  \label{fig:acc_cm}
\end{figure}

For the results below, the feedback agent (FL) was disabled due to data sensitivity.
The preliminary results appear in Table~\ref{tab:five-one}, and the corresponding
confusion matrix is provided in Fig.~\ref{fig:acc_cm}.

\section*{Discussion \& Conclusion} 
The ORCHID framework improves classification reliability, transparency, and reproducibility through evidence-based policy-aware decision-making. Using RAG, each classification is grounded in traceable citations, ensuring verifiable reasoning. Its hybrid retrieval mechanism integrates domain-specific regulatory corpora, ITAR/USML, NRC, CCL, EAR99, for policy compliance, while a human-in-the-loop design incorporates expert feedback to refine performance and prevent recurring errors. ORCHID’s modular, agentic architecture supports scalability and reproducibility, and its single-click interface streamlines the decision process for efficient, auditable outcomes.


The current implementation of ORCHID faces several practical limitations. Its performance depends on curated policy corpora, making it sensitive to coverage gaps and drift when source texts become outdated. 

\begin{table}[!b]
  \centering
  \caption{Comprehensive preliminary accuracy results.}
  \label{tab:five-one}
  \begin{tabular}{|c|c|c|c|c|c|}
    \hline
    USML & NRC & CCL & EAR99 & Weighted Avg. & Binary Acc \\
    \hline
    88\% & 90\% & 56\% & 40\% & 63.12\% & 70.37\% \\
    \hline
  \end{tabular}
\end{table}

Boundary ambiguity persists in fine-grained classifications, particularly in distinguishing CCL and EAR99 items, where validator calibration remains an ongoing effort. The framework currently supports only English text and does not process multimodal inputs such as images or technical specification sheets. In addition, the quality of retrieval and classification is reduced with sparse or poorly written descriptions, and the ``no-description'' mode exhibits reduced classification reliability. ORCHID provides decision support but does not constitute legal or regulatory advice, and final determinations must be made by qualified reviewers.

\section*{Acknowledgments}
This research is sponsored by the Office of the Laboratory Director, Oak Ridge National Laboratory's Operational Excellence Initiatives, which is supported by the United
States Department of Energy (DOE)’s Office of Science under Contract No. DE-AC05-00OR22725. We thank the ORNL Operations and HRP SME team for their valuable expertise, feedback, and domain insights throughout this project.

\bibliographystyle{unsrt}      

\bibliography{ref}




\end{document}